\documentclass[conference]{IEEEtran}
\IEEEoverridecommandlockouts
\usepackage{graphicx}
\usepackage{xcolor}
\ifCLASSINFOpdf
\else
\fi
\usepackage{url}

\usepackage{cite}
\usepackage{hyperref}
\usepackage{amsmath,amssymb,amsfonts}
\usepackage{algorithmic}
\usepackage{graphicx}
\usepackage{textcomp}
\usepackage{xcolor}
\usepackage{bm} 
\usepackage{float} 

\newcommand{\action}{\mathcal{A}}
\newcommand{\state}{\mathcal{S}}
\newcommand{\obs}{\mathcal{O}}

\newcommand{\bolda}{\bm{a}}
\newcommand{\boldo}{\bm{o}}
\newcommand{\boldpi}{\bm{\pi}}
\newcommand{\ratio}{\frac{\policyF(\bolda_t| s_t)}{\boldpi_{\thetaOld}(\bolda_t| s_t)}}
\newcommand{\ratioI}{\frac{\policyF(a^i_t| o^i_t)}{\boldpi_{\thetaOld}(a^i_t| o^i_t)}}
\newcommand{\clip}{\text{clip}}

\newcommand{\phiOld}{\phi_{\text{old}}}
\newcommand{\thetaOld}{\theta_{\text{old}}}
\newcommand{\loss}{\mathcal{L}}

\newcommand{\policyF}{{\boldpi_{\theta}}}
\newcommand{\valueF}{{V_{\phi}}}
\newcommand{\advantage}{\Hat{A}_t}
\newcommand{\Vhat}{\Hat{V}}

\newcommand{\concat}{\text{Concat}}
\newcommand{\head}{\text{head}}

\newcommand{\feedforward}{\text{FF}}
\newcommand{\ff}{\feedforward}
\newcommand{\Xhead}{\mathcal{X}}

\newcommand{\NATURAL}{\mathbb{N}}
\newcommand{\expect}{\mathop{\mathbb{E}}}

\begin{document}
%
\title{
    Resolve Highway Conflict in Multi-Autonomous Vehicle Controls with Local State Attention
}

\makeatletter
\newcommand{\linebreakand}{%
  \end{@IEEEauthorhalign}
  \hfill\mbox{}\par
  \mbox{}\hfill\begin{@IEEEauthorhalign}
}
\makeatother

\author{\IEEEauthorblockN{Xuan Duy Ta}
\IEEEauthorblockA{VNU University of Engineering and Technology \\ Hanoi, Vietnam}
\and
\IEEEauthorblockN{Bang Giang Le}
\IEEEauthorblockA{VNU University of Engineering and Technology \\ Hanoi, Vietnam}
\linebreakand
\IEEEauthorblockN{Thanh Ha Le}
\IEEEauthorblockA{VNU University of Engineering and Technology \\ Hanoi, Vietnam}
\and
\IEEEauthorblockN{Viet Cuong Ta$^*$\thanks{$^*$Corresponding Author}}
\IEEEauthorblockA{VNU University of Engineering and Technology \\ Hanoi, Vietnam}}


%


\maketitle

\begin{abstract}
In mixed-traffic environments, autonomous vehicles must adapt to human-controlled vehicles and other unusual driving situations.
This setting can be framed as a multi-agent reinforcement learning (MARL) environment with full cooperative reward among the autonomous vehicles.
While methods such as Multi-agent Proximal Policy Optimization can be effective in training MARL tasks, they often fail to resolve local conflict between agents and are unable to generalize to stochastic events.
In this paper, we propose a Local State Attention module to assist the input state representation.
By relying on the self-attention operator, the module is expected to compress the essential information of nearby agents to resolve the conflict in traffic situations.
Utilizing a simulated highway merging scenario with the priority vehicle as the unexpected event, our approach is able to prioritize other vehicles' information to manage the merging process. The results demonstrate significant improvements in merging efficiency compared to popular baselines, especially in high-density traffic settings.
\end{abstract}


%
\IEEEpeerreviewmaketitle

\section{Introduction}

Autonomous vehicle (AV) technologies empower vehicles to undertake various driving tasks without human intervention. Among many challenging driving scenarios, highway merging stands out as one of the most difficult for AVs \cite{Daiheng2005mergebottleneck, Leclercq2011capacityDropsAtMerges}, while also being a primary source of traffic bottlenecks \cite{Malikopoulos2013proofonmergeisbottleneck}. This complexity is further exacerbated in mixed-traffic environments that include both AVs and human-driven vehicles, as well as the presence of priority vehicles.
Resolving the conflict between speed and safety conditions remains an insurmountable challenge. The requirement for AVs to generalize their behavior across a wide variety of situations magnifies its sophistication, with the challenge intensifying as traffic density increases.

The integration of deep reinforcement learning (RL) techniques into various connected and autonomous vehicle scenarios has been extensively researched.
In these applications, state representation plays a pivotal role in the efficacy of RL algorithms.
The attention mechanism \cite{vaswani2023attention} capable of identifying inter-dependencies among inputs in a self-supervised manner, proves to be highly effective for these challenges. Specifically, \cite{Liu2024SceneRepresentationLearning_transformer} employs transformers with off-policy RL for general urban scenarios, while \cite{Chen2024attentionDRLhighwaysafety, Wang2022attentionDRLhighwaylanechange} utilizes the attention mechanism to enhance RL performance in highway environments.
In the context of multi-agent reinforcement learning (MARL), the authors of \cite{schester2021ramp_MAQlearning} employed multi-agent Q-learning, \cite{Zhou2022ramp_MADDPG} utilized multi-agent Deep Deterministic Policy Gradient, both studies tackling highway on-ramp merging.
Other multi-agent reinforcement learning approaches have been developed for traffic control \cite{ Zhang2023ramp_IPPO}.

Employing multi-agent systems for traffic management can significantly elevate the efficacy of cooperative behavior, surpassing the capabilities of single-agent approaches.

From a multi-agent reinforcement learning (MARL) perspective, QMIX \cite{ rashid2020monotonic} is a pioneering method for fully cooperative scenarios. It extends value-based deep Q-learning by decomposing the global Q-function into individual agent contributions while ensuring monotonicity constraints. Other notable value-based methods include MADDPG \cite{lowe2017multi} and VDN \cite{sunehag2017value}.
Compared with value-based methods, policy-based methods often achieve faster training speeds due to their ability to leverage gradient-based optimization directly on policies. Popular policy-based optimization methods for MARL include Independent PPO \cite{dewitt2020ippo} and Multi-Agent PPO (MAPPO) \cite{yu2022surprising}.
Among these, MAPPO is considered the leading baseline for homogeneous agent scenarios due to its superior performance across various benchmarks.
However, mitigating the speed-safety dichotomy in traffic control presents a formidable challenge in cooperative MARL for these approaches. Balancing the trade-off between high-speed efficiency and ensuring safety becomes exceedingly complex in dynamic and unpredictable traffic environments, where AVs must adapt to varying traffic densities, the behavior of human-driven vehicles, and the presence of priority vehicles. Ensuring that AVs can generalize their decision-making processes across such a wide range of conditions while consistently prioritizing safety without significantly compromising speed requires sophisticated algorithms and extensive training.

In this work, we aim to tackle multi-AV traffic control by proposing an attention-based state encoder.
Given the important role of local observations in ensuring the trade-off between speed and safety for each AV in multi-agent settings, our Local State Attention mechanism is designed to learn more efficient representations which could reduce stochasticity in high-density traffic situations.
Our extension is integrated into the popular MAPPO algorithm and verified in the settings of highway merging.
By varying the traffic density and introducing unexpected merging conflicts via priority vehicles, our proposed approach is able to outperform other baselines in 5 testing scenarios. 
Further testing results highlight that, by leveraging Local State Attention, the joint policy is more robust and can generalize better in complex traffic situations.

\section{Related Work and Backgrounds}
\textbf{MARL methods.}
Two popular approaches are value-based and policy gradient-based multi-agent methods. 
In valued-based methods, 
\cite{lowe2017multi} introduced MADDPG by extending DDPG for multi-agents with a centralized critic.
QMIX \cite{rashid2020monotonic} extends the value decomposition to the more flexible combination of value functions based on a monotonicity constraint.
MARL agents typically suffer from training instability due to the non-stationary of other exploring agents \cite{kuba2021settling}, as well as partial observability, which violates the Markov assumption. 
Policy gradient methods, such as PPO \cite{schulman2017proximal}, are among the approaches that do not rely on the Markov assumption.
MAPPO \cite{yu2022surprising}, which is an extension from the PPO, is currently the most popular on-policy MARL method which utilizes trust region optimization under the centralized training framework.  

\textbf{Attention in MARL.} 
Attention is extensively used in communication MARL, for example in inter-agent communication \cite{das2019tarmac}.
Besides, the agents can rely on the attention operator for value decomposition as proposed in \cite{yang2020qatten}. 
Additionally, attention is also used to simplify the interaction of agents in large-scale partially observable environments \cite{zhang2022multi} or to learn an implicit coordination graph for improving coordination \cite{li2020deep}.

\textbf{MARL framework} We model the mixed traffic environment between AVs and human-control vehicles as a decentralized partially observable Markov decision process (Dec-POMDP) \cite{Oliehoek2016decPOMDPs}, defined by the tuple $\langle \state, \action, \obs, R, P, \omega, n, \gamma \rangle$. Here, $n$ is the number of agents. $\state, \action$ is the shared state and action space of all agents, respectively. The initial state at timestep 0, $s_0$, is drawn from distribution $\omega$. The discount factor $\gamma\in [0;1)$ is used for controlling the future rewards.

At any timestep $t$ with state $s_t\in\state$, each agent $i$ can only observe its local observation $o^i_t = \obs(s_t;i)$. The joint observation of all agents at timestep $t$ is $\boldo_t = \langle o^1_t, o^2_t,... o^n_t \rangle$. Each agent using only its own history of observations and actions up to this point, $\tau^i_t$, chooses an action with a decentralised policy $a^i_t \sim \pi^i(\cdot|\tau^i_t)$. After executing the joint action $\bolda_t = \langle a^1_t, a^2_t, ... a^n_t\rangle \in \action^n$, agents all receive the common scalar reward $r_t = R(s_t,\bolda)$ and the next state $s_{t+1}$ is selected following the distribution $P(s_t, \bolda)$.
Our setting for the mixed traffic control problem is the fully cooperative reward which the reward of agent $i$ is the same and equals $r^i_t = r_t$.

The objective is to learn the joint policy $\boldpi_\theta = \langle \pi^1, \pi^2, ..., \pi^n \rangle$ parameterized by $\theta$ that maximizes the expected discount reward $J(\boldpi) = \expect[\sum_{i=0}^{\infty}\gamma^t r_{t}]$.

\section{Methods}
\begin{figure*}[htbp]
    \centering
    \includegraphics[width=0.8\textwidth]{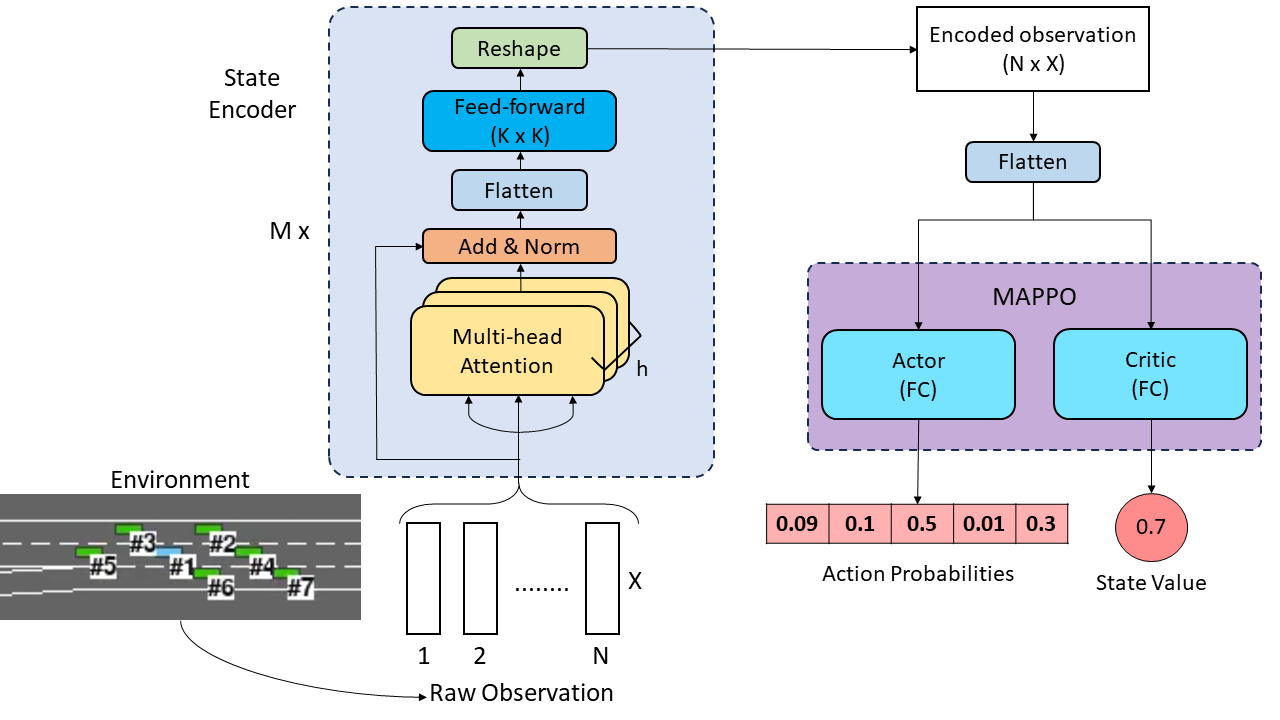}
    \caption{Our approach for aggregating the information by employing a Local State Attention for the Actor and Critic of MAPPO. The input observation of Agent 1 (blue) is concatenated from a list of features $X$ from $N$ nearby agents.}
    \label{fig:control-approach-marl}
\end{figure*}
For controlling multi-AVs in traffic situations, our work employs MAPPO, a direct extension of the PPO \cite{schulman2017proximal} algorithm with the clipping mechanism.
The underlying theory of PPO is the Trust Region Policy Optimization Framework \cite{schulman2017trpo} which constrains the policy update step size.
We first introduce the key approach of MAPPO and our state encoder for resolving the conflict in multi-AV control.

\subsection{Multi-agent Proximal Policy Optimization}
Proximal Policy Optimization (PPO) is a family of first-order algorithms that emulates the monotonic improvement of trust-region optimization by approximating the Kullback–Leibler divergence through clipping the policy ratio. Denoting $\rho_t(\theta) = \ratio$ and overloading the notation $\clip(x,y) = \clip(x, y-\epsilon, y+\epsilon)$, the policy loss is:
\begin{equation}
    \loss(\theta) = \expect_{s_t,\bolda_t}\left[ \min\left(\rho_t(\theta) \advantage, \clip(\rho_t(\theta),1) \advantage\right) \right]
\end{equation}

Multi-agent PPO (MAPPO) \cite{yu2022surprising} extend PPO to multi-agent's Dec-POMDP environments following the \textit{centralised training with decentralised execution (CTDE)}\cite{Oliehoek2016decPOMDPs} training framework. We generate the advantage function estimate for each agent $\advantage^i$ with the help of an observation-based critic $\valueF$ parameterised by $\phi$ using Generalized Advantage Estimation following the equations:
\begin{gather}
    \advantage^i = \sum_{l=0}^{k} (\gamma\lambda)^l \delta_{t+l}, \\
    \delta_t = r_t + \gamma\valueF(\boldo_{t+1}) - \valueF(\boldo_t),
\end{gather}
where $\lambda$ is a hyper-parameter, $\boldo_t$ is the joint observation of all agents at timestep $t$. We consider the case where parameter sharing is employed ($\policyF = \pi^1 = \pi^2 = ... = \pi^n$). The algorithms' policy loss for each agent $i$ is:
\begin{gather}
    \loss^i(\theta) = \expect_{o^i_t, a^i_t}\left[
        \min\left(
            \rho^i_t(\theta) \advantage^i,
            \clip(\rho^i_t(\theta),1) \advantage^i
        \right) 
    \right], \\
    \rho^i_t(\theta) = \ratioI,
\end{gather}

The critic loss is also clipped, as purposed by \cite{schulman2018gae}:
\begin{gather}
    \loss^i(\phi) = \expect_{\boldo_t} \left[
        \max\left\{
            \left(\valueF(\boldo_t) - \Vhat^i_t\right)^2,
            \left(\clip V - \Vhat^i_t\right)^2
        \right\}
    \right], \\
    \Vhat^i_t = \advantage^i + \valueF(\boldo_t), \\
    \clip V = V_{\phiOld}(\boldo_t) + \clip(\valueF(\boldo_t) - V_{\phiOld}(\boldo_t), 0),
\end{gather}
where $\phiOld$ is the critic parameters before the update.

The overall loss is formulated as follows:
\begin{equation}
    \loss(\theta, \phi) = \sum_{i=1}^n { \loss^i(\theta) - \beta_1 \loss^i(\phi) + \beta_2 \mathcal{H}(\policyF) },
\end{equation}
where $\mathcal{H}(\policyF)$ is the entropy of the actor and $\beta_1,\beta_2$ are positive weighting coefficients for the critic and the entropy.

\subsection{Local State Attention for Unifying Agents' Observation}

In practical multi-AV environments, the features each AV observes are well-defined. Consider the general case where each agent observes $X$ features of at most $N$ closest vehicles (including itself) that are within a radius around it ($o^i_t$ is a $N\times X$ matrix for all agent $i$, timestep $t$).
We can consider this as a local state representation of a specific agent.
For resolving conflict between AVs and other vehicles' actions, it is possible to rely on the local state to make proper actions.
However, the issues arose mainly with distinguishing between non-existent vehicles and vehicles with similar features to itself along with the high variance associated with estimating model parameters.
In addition to that, a MAPPO state encoder should have the capability to weigh different features and vehicles when considering its next actions.
For example, the position and velocity of a vehicle can help predict its future position and the distance and relative speed between two vehicles can help determine the need to yield the agent's right of way.
Therefore, we employ a Local State Attention (LSA) as a state encoder to more effectively extract the latent representation of an agent's surrounding environment in traffic situations.
The proposed state encoder is illustrated in Fig.\ref{fig:control-approach-marl}.

The state encoder contains $M$ blocks of multi-head self-attention.
The input of the state encoder is the raw local observation of $N$ closest vehicles combined into a $N\times X$ matrix.
The output representation is flattened and then directly fed to both the Actor and Critic networks.

\subsubsection{Multi-Head Self Attention} 
Similar to the Transformer \cite{vaswani2023attention}, we use scaled dot-product attention for aggregating the local information.
Given matrices $Q,K$ and $V$ all of shape $\mathcal{N}\times\Xhead$, it multiplies the queries $Q$ and the transpose of the keys $K^{T}$, scale by $\frac{1}{\sqrt{\Xhead}}$ before applying a non-linear function $\sigma$ and then multiply with the value matrix $V$. Formally, the self-attention operator is defined as:
\begin{equation}
    \text{Attention}(Q,K,V) = \sigma\left(\frac{QK^{T}}{\sqrt{\Xhead}}\right)V
\end{equation}
where $\sigma$ is usually the softmax function.

Instead of performing a single attention function, it is beneficial to linearly project the queries, keys and values $h$ times. Each of these \textit{attention heads} will perform the aforementioned attention function in parallel, the results are concatenated and projected once more before outputting. Specifically:
\begin{align}
    \text{Multi-Head}(Q,K,V) = \concat(\head_1,...,\head_h)W^O, \\
    \head_i = \text{Attention}(QW^Q_i, KW^K_i, VW^V_i),
\end{align}
Here, projection matrices $W^Q_i, W^K_i, W^V_i$ are $X \times \Xhead$, $W^O$'s shape is $X\times X$. Inputs $Q = K = V = o^i_t$ and have dimension $N\times X$. Each attention head is computed with $\mathcal{N} = N$ and $\Xhead = X / h$. $h$ is to be chosen such that $\Xhead \in \NATURAL$, usually each attention head focuses on a different feature aspect.

\subsubsection{Feed-forward Network}
The fully connected feed-forward network in each of our State Encoder stack has input and output dimensions $K = N\cdot X$. We flatten the multi-head attention's output as well as reshape the feed-forward network's output back to $N\times X$. Our feed-forward network is a linear matrix multiplication:
\begin{equation}
    \ff(\mathcal{Y}) = \mathcal{Y} W^{\ff}
\end{equation}
where $W^{\ff}$ is a $K\times K$ matrix.
It is expected that our LSA is capable of modeling the interaction of input observation $o^i_t$.
The local features in $o^i_t$, such as the position and velocity of the vehicles, can complement each other and reduce the stochasticity of the environment, especially when the number of agents increases.

\begin{figure}[htbp]
    \center{
    \includegraphics[width=0.5\textwidth]{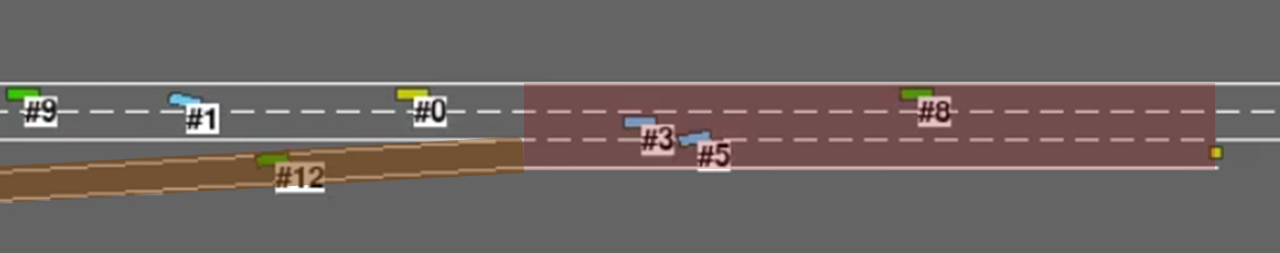}
    \caption{
    Partial screenshot of the simulation in scenario 5 showing the merge lane (light brown) and highway; including the Priority Vehicle (yellow), Connected and Autonomous Vehicles (blue) and Human-Driven Vehicles (green). The major conflict happens in the red area.
    In this example, Agent 3 wants to maintain velocity while keeping its left lane open for the rapidly approaching priority vehicle. 
    }
    \label{fig:sample-simulation}
    }
\end{figure}
\begin{table}[h]
    \centering
    \caption{The number of each type of vehicle, traffic densities and reward of MAPPO baseline of 5 tested scenarios}
    \label{tab:traffic_densities}
    \begin{tabular}{|c|c|c|c|c|}
        \hline
        \textbf{Task} & \textbf{CAVs} & \textbf{HDVs} & \textbf{Traffic Density} & \textbf{Reward} \\\hline
        Scenario 1 & 2 & 4 & Light  & -25 \\\hline
        Scenario 2 & 3 & 3 & Light  & -120 \\\hline
        Scenario 3 & 4 & 2 & Medium & -140 \\\hline
        Scenario 4 & 4 & 4 & Medium & -150 \\\hline
        Scenario 5 & 6 & 6 & Heavy  & -160 \\\hline
    \end{tabular}
\end{table}

\section{Experiment}

\begin{figure*}[htbp]
    \center{
    \includegraphics[width=1.0\textwidth]{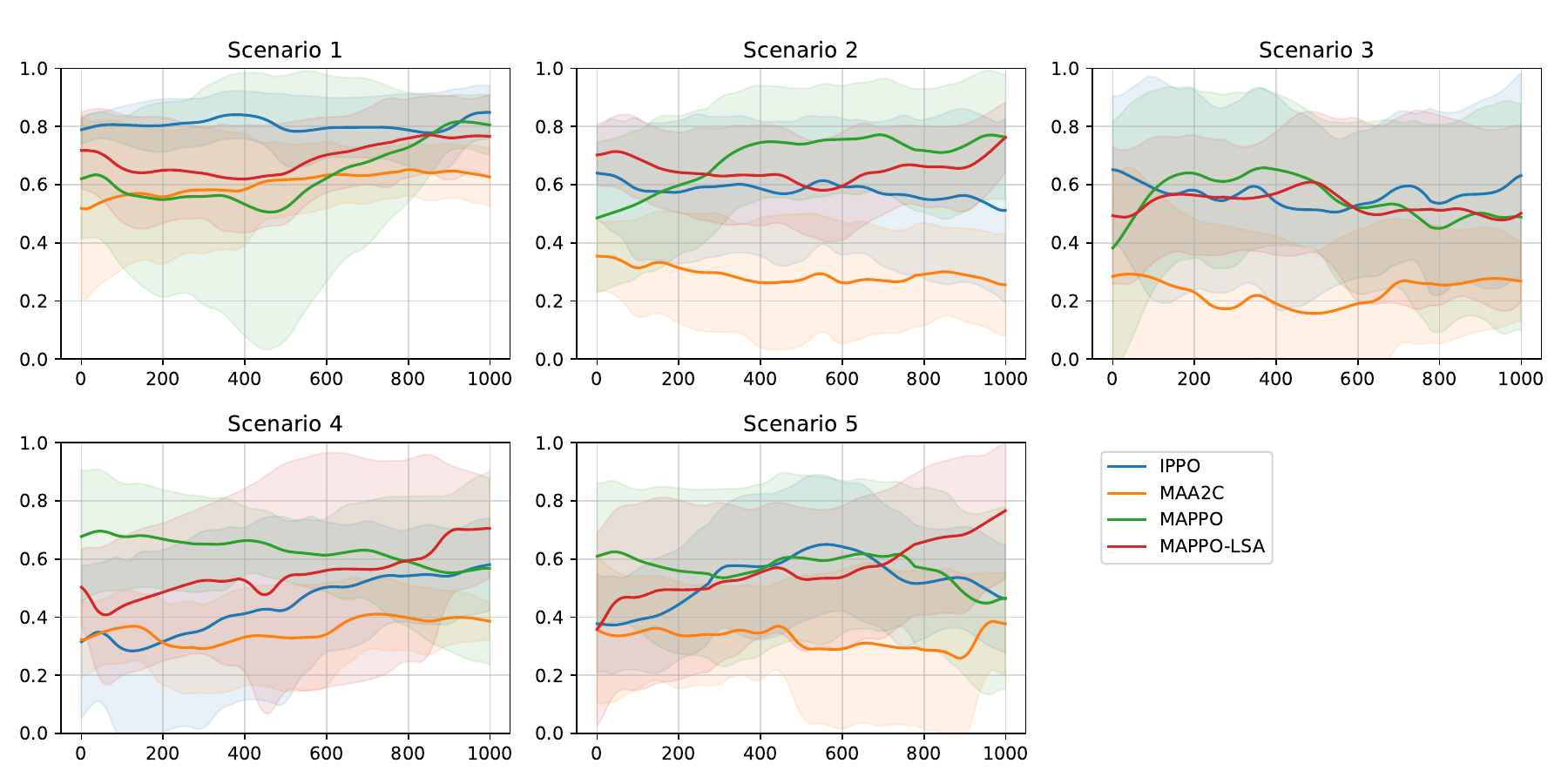}
    \caption{
        Performance comparison in different scenarios for 1000 training epochs. The rewards are normalized to [0, 1] for better comparison.
    }
    \label{fig:rewards}
    }
\end{figure*}

\subsection{Experiment Setup}
We evaluate our proposed method by expanding the multi-agent highway environment introduced by \cite{chen2023deep}, simulating a mixed-traffic highway on-ramp merging scenario with priority vehicles' presence.
Figure \ref{fig:sample-simulation} presents a snapshot of our simulation with 6 AVs. The agents are colored blue.
We model the environment as a 2-lane highway with connected and autonomous vehicles (CAVs), human-driven vehicles (HDVs) and priority vehicles (PVs).
The HDVs and PVs are controlled by the simulator \cite{Leurent2018highwayenv}. 
Without loss of generalization, we only consider the presence of a single priority vehicle.
To create traffic conflicts, this priority vehicle is set to prefer traveling at double the speed of the HDVs and to maintain its right of way. For the other vehicles, CAVs and HDVs, we adjust their numbers to vary the traffic density.

The CAVs represent the controlled agents which will be learned by the multi-agent policies.
The action space $\action$ is the set of 5 discrete high-level control options, including \textit{turn left, turn right, cruising, speed up} and \textit{slow down}. 
The observation of agent $i$ at timestep $t$, $o^i_t$, is a matrix of dimension $N\times X$, where $N$ is the maximum number of observable vehicles - including itself, and $X=6$ is the amount of features to be observed, comprising of $[b,p, x,y, vx,vy]$, specifically:
\begin{itemize}
    \item $b, p$: Binary variable denoting whether a vehicle is observed
    and whether the observed vehicle is the priority vehicle, respectively.
    \item $x, y$: Longitudinal and Lateral position of the observed vehicle, relative to the agent. 
    \item $vx, vy$: Longitudinal and lateral velocity of the observed vehicle, relative to the agent. 
\end{itemize}

We employ a reward function, which heavily punishes unsafe driving acts such as stalling, crashing or weaving, designed to encourage agents to navigate safely through the merging area while maintaining a proper speed and making way for the priority vehicle. For comparison purposes, we normalized the reward to range $[0,1]$.

We compare our method, MAPPO-LSA, with three other baselines for 1000 epochs across 3 different seeds, which are the popular on-policy algorithms: MAPPO \cite{yu2022surprising}, IPPO \cite{dewitt2020ippo} and MAA2C - a direct extension of the A2C \cite{mnih2016asynchronous} algorithm to multi-agent learning.

The actor and critic are separate networks featuring three fully connected layers with dimensions $[K, 256, 256]$ with Tanh activation. The critic then outputs a single number, while the actor's output layer has 5 units followed by another Tanh activation.
For the LSA encoder, we set the number $M = 1$; the number of heads, $h = X/2$, each head focusing on a different observation subset that shares the same units.

We consider various levels of traffic densities, characterized by different numbers of CAVs and HDVs. The environments are numbered from 1 to 5, with 1 being the easiest and 5 being the hardest.
The traffic density rises with an increased number of CAVs and number of total vehicles, as shown in Table \ref{tab:traffic_densities}.
MAPPO's unscaled average reward after 1000 training epochs is also provided to help put into context the relative environment difficulty.
We set the maximum number of observable vehicles to $N = 4$ for the two light traffic density settings, and $N = 6$ for the remaining scenarios.

\subsection{Results}
The results of 5 tested scenarios are illustrated in Fig.\ref{fig:rewards}.
MAPPO-LSA demonstrates superior performance in high-density traffic environments while maintaining comparable results to baselines in other settings. Notably, in the highest density scenarios - scenarios 4 and 5 - MAPPO-LSA displays effective learning and continuous improvement, whereas the baselines struggle to establish a functional policy. This highlights how the State Encoder assists MAPPO in maintaining robustness and adaptability in more challenging and complex traffic conditions. In less dense environments, MAPPO-LSA performs on par with other baseline models, showing consistent performance across different settings. For the other baselines, MAPPO and IPPO have a comparable performance which is better than MAA2C,

We further conduct an ablation study to investigate the two important aspects of the LSA module: the maximum number of vehicles an agent can use at a time $N$ and the quality and quantity of input features for each observed vehicle $X$.
We run the experiments through light, medium and heavy traffic settings, specifically Scenario 2, 4 and 5.

Testing different maximum observable vehicle quantities $N = 2, 4,$ and $6$ reveals distinct performance patterns (Fig.\ref{fig:ablation-rewards-N}).
In light traffic environments, a lower $N$ is preferable, yielding better results.
However, as traffic density increases, a higher $N$ becomes advantageous. Specifically, in denser settings, larger $N$ values allow for better performance and more effective learning. This trend highlights the need to adjust $N$ based on traffic density to optimize the model's performance. Alternatively, modifying the observation processing method, treating observations as variable-length sequences with max length $N$ instead of a fixed-size $N\times X$ matrix, can be beneficial.

For revealing the important feature in controlling the AVs, we also alternate the available features which are inputted into the LSA module: \textit{No position} - the position information $(x,y)$ is masked; \textit{No presence or priority} - the presence and priority are masked; \textit{No velocity} - the velocity information $(vx,vy)$ is omitted. 
We also add a \textit{Add angles} configuration, which is a pair $(\cos(\zeta), \sin(\zeta))$ given $\zeta$ is the angle that the vehicle makes with the highway.
The results are illustrated in Fig.\ref{fig:ablation-rewards-X}.
Our findings indicate that in low and medium-density scenarios, agents can effectively learn policies despite having incomplete information. However, dense traffic settings reveal that the quality and quantity of information are crucial. Specifically, the absence of neighboring vehicles' velocity significantly impedes policy learning, while excessive information, such as vehicle angles, introduces noise and distracts the agent.
Interestingly, the lack of precise relative positions of vehicles does not affect performance.

\begin{figure*}
    \center{
    \includegraphics[width=0.9\textwidth]{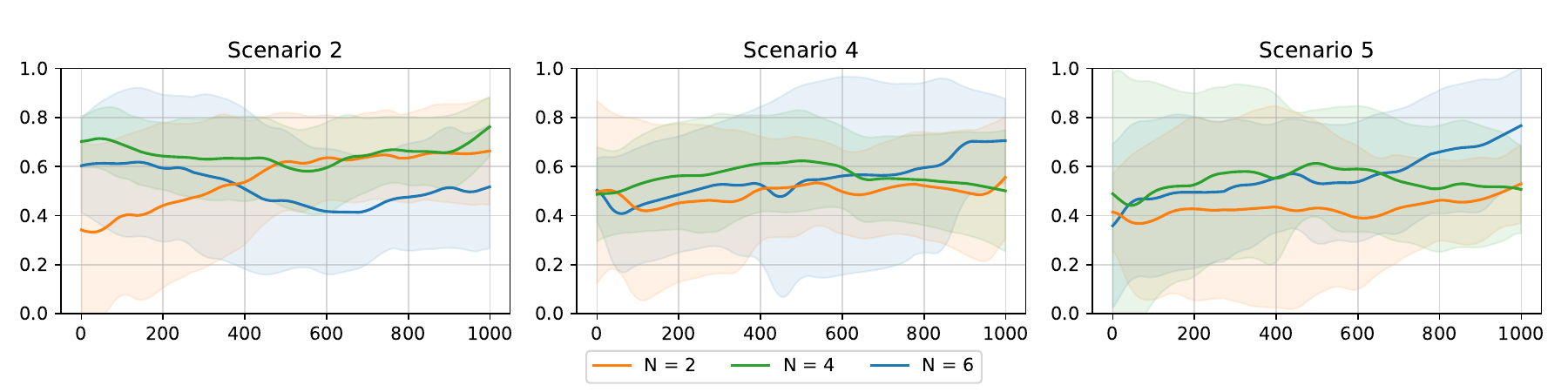}
    \caption{Performance comparison between different maximum observable vehicles $N$ of MAPPO-LSA}
    \label{fig:ablation-rewards-N}
    }
\end{figure*}

\begin{figure*}
    \center{
    \includegraphics[width=0.9\textwidth]{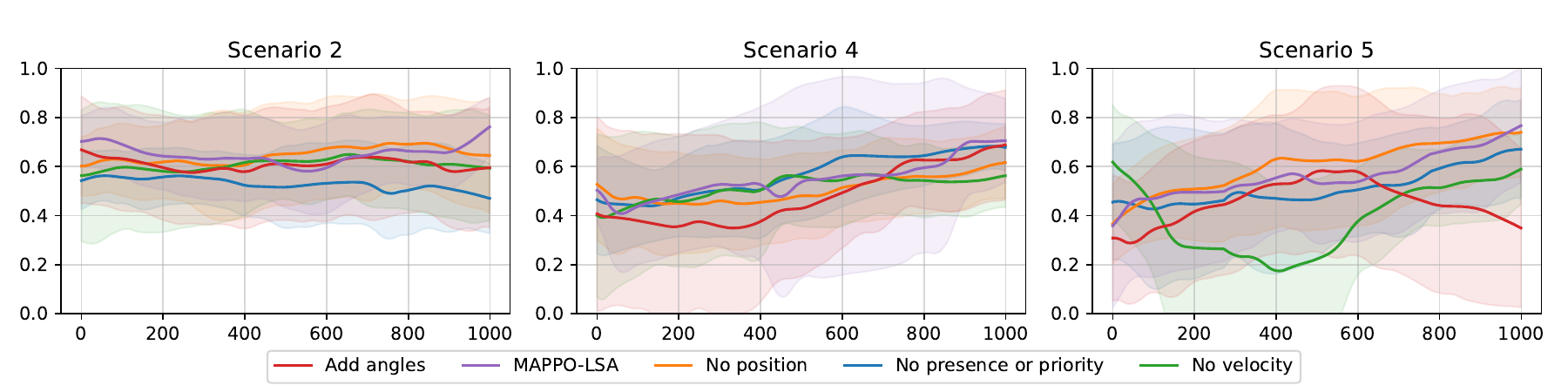}
    \caption{Performance comparison between different input observation features of MAPPO-LSA}
    \label{fig:ablation-rewards-X}
    }
\end{figure*}

\section{Conclusion}
Our attention-based deep MARL approach demonstrated significant improvements in merging efficiency and safety in complex mixed-traffic environments. The enhanced performance of our approach can be attributed to the attention mechanism's effective local state representation. Compared to traditional baselines, MAPPO-LSA results in smoother and more efficient on-ramp merging even in the presence of PVs. The method was evaluated across a wide range of traffic volumes, demonstrating training stability in high-density traffic, where common baselines fail to learn a proper policy.


\section*{Acknowledgment}
This material is based upon work supported by the Air Force Office of Scientific Research under award number FA2386-24-1-4012.




\bibliographystyle{IEEEtran}
\bibliography{main}

\end{document}